\documentclass[letterpaper,11pt]{article}

\usepackage[latin1]{inputenc}
\usepackage[T1]{fontenc}
\usepackage{amsfonts}
\usepackage{amssymb}
\usepackage{amsmath} 
\usepackage{amsthm}
\usepackage{graphicx} 
\usepackage{subfigure} 
\usepackage{multirow}

\newcommand{\Xcal}{\mathcal{X}}
\newcommand{\Pcal}{\mathcal{P}}
\newcommand{\Ucal}{\mathcal{U}}
\newcommand{\RR}{\mathbb{R}}
\newcommand{\PP}{\mathbb{P}}
\newcommand{\OMIT}[1]{}

\makeatletter
\def\captionof#1#2{{\def\@captype{#1}#2}}
\makeatother

\usepackage{url}
\usepackage{rotating}
\usepackage{algorithm}
\usepackage{algorithmic}
\usepackage{amsmath,amsfonts,amssymb}
\usepackage{natbib}

\usepackage[margin=1in]{geometry}

\begin{document}

\title{A bagging SVM to learn from positive and unlabeled examples}

\author{
Fantine Mordelet\\
Institut Curie, Paris, F-75248 France\\
INSERM, U900, Paris, F-75248 France\\
Ecole des Mines de Paris F-77300 France\\
\texttt{fantine.mordelet@mines-paristech.fr}
\and
Jean-Philippe Vert\\
Institut Curie, Paris, F-75248 France\\
INSERM, U900, Paris, F-75248 France\\
Ecole des Mines de Paris F-77300 France\\
\texttt{jean-philippe.vert@mines-paristech.fr}
}

\maketitle

\begin{abstract}
We consider the problem of learning a binary classifier from a training set of positive and unlabeled examples, both in the inductive and in the transductive setting. This problem, often referred to as \emph{PU learning}, differs from the standard supervised classification problem by the lack of negative examples in the training set. It corresponds to an ubiquitous situation in many applications such as information retrieval or gene ranking, when we have identified a set of data of interest sharing a particular property, and we wish to automatically retrieve additional data sharing the same property among a large and easily available pool of unlabeled data. We propose a conceptually simple method, akin to bagging, to approach both inductive and transductive PU learning problems, by converting them into series of supervised binary classification problems discriminating the known positive examples from random subsamples of the unlabeled set. We empirically demonstrate the relevance of the method on simulated and real data, where it performs at least as well as existing methods while being faster.
\end{abstract}

\section{Introduction}

In many applications, such as information retrieval or gene ranking, one is given a finite set of data of interest sharing  a particular property, and wishes to find other data sharing the same property. In information retrieval, for example, the finite set can be a user query, or a set of documents known to belong to a specific category, and the goal is to scan a large database of documents to identify new documents related to the query or belonging to the same category. In gene ranking, the query is a finite list of genes known to have a given function or to be associated to a given disease, and the goal is to identify new genes sharing the same property \citep{Aerts2006Gene}. In fact this setting is ubiquitous in many applications where identifying a data of interest is difficult or expensive, e.g., because human intervention is necessary or expensive experiments are needed, while unlabeled data can be easily collected. In such cases there is a clear opportunity to alleviate the burden and cost of interesting data identification with the help of machine learning techniques.

More formally, let us assign a binary label to each possible data: positive ($+1$) for data of interest, negative ($-1$) for other data. Unlabeled data are data for which we do not know whether they are interesting or not. Denoting $\Xcal$ the set of data, we assume that the ``query'' is a finite set of data $\Pcal=\{x_1,\ldots,x_m\} \subset \Xcal$ with positive labels, and we further assume that we have access to a (possibly large) set $\Ucal = \{x_{m+1},\ldots,x_n\}$ of unlabeled data. Our goal is to learn, from $\Pcal$ and $\Ucal$, a way to identify new data with positive labels, a problem often referred to as \emph{PU learning}. More precisely we make a distinction between two flavors of PU learning:
\begin{itemize}
\item \emph{Inductive PU learning}, where the goal is to learn from $\Pcal$ and $\Ucal$ a function $f:\Xcal\rightarrow \RR$ able to associate a score or probability to be positive $f(x)$ to any data $x\in\Xcal$.
This may typically be the case in an image or document classification system, where a subset of the web is used as unlabeled set $\Ucal$ to train the system, which must then be able to scan any new image or document.
\item \emph{Transductive PU learning}, where the goal is estimate a scoring function $s:\Ucal\rightarrow\RR$ from $\Pcal$ and $\Ucal$, i.e., where we are just interested is finding positive data in the set $\Ucal$. This is typically the case in the disease gene ranking application, where the full set of human genes is known during training and split between known disease genes $\Pcal$ and the rest of the genome $\Ucal$. In that case we are only interested in finding new disease genes in $\Ucal$. 
\end{itemize}

Several methods for PU learning, reviewed in Section \ref{sec:related} below, reduce the problem to a binary classification problem where we learn to discriminate $\Pcal$ from $\Ucal$. This can be theoretically justified, at least asymptotically, since the log-ratio between the conditional distributions of positive and unlabeled examples is monotonically increasing with the log-ratio of positive and negative examples \citep{Elkan2008Learning,Scott2009Novelty}, and has given rise to state-of-the-art methods such as \emph{biased support vector machine (biased SVM)} \citep{Liu2003Building} or weighted logistic regression \citep{Lee2003Learning}. Although this reduction suggests that virtually any method for (weighted) supervised binary classification can be used to solve PU learning problems, we put forward in this paper that some methods may be more adapted than others in a non-asymptotic setting, due to the particular structure of the unlabeled class. In particular, we investigate the relevance of methods 
based on aggregating classifiers trained on artificially perturbed training sets, in the spirit of bagging \citep{Breiman1996Bagging}. Such methods are known to be relevant to improve the performance of unstable classifiers, a situation which, we propose, may occur particularly in PU learning. Indeed, in addition to the usual instability of learning algorithms confronted to a finite-size training sets, the content of a random subsample of unlabeled data in positive and negative examples is likely to strongly affect the classifier, since the contamination of $\Ucal$ in positive examples makes the problem more difficult. Variations in the contamination rate of $\Ucal$ may thus have an important impact on the trained classifier, a situation which bagging-like classifiers may benefit from.

Based on this idea, we propose a general and simple scheme for inductive PU learning, akin to an asymetric form of bagging for supervised binary classification. The method, which we call \emph{bagging SVM}, consists in aggregating classifiers trained to discriminate $\Pcal$ from a small random subsample of $\Ucal$, where the size of the random sample plays a specific role. This method can naturally be adapted to the transductive PU learning framework. We demonstrate on simulated and real data that bagging SVM performs at least as well as existing methods for PU learning, while being often faster in particular when $|\Pcal | << | \Ucal |$.

This paper is organized as follows. After reviewing related work in Section \ref{sec:related}, we present the bagging SVM for inductive PU learning in Section \ref{sec:inductive}, and its extension to transductive PU learning in Section \ref{sec:transductive}. Experimental results are presented in \ref{sec:res}, followed by a Discussion in Section \ref{sec:discussion}.

\OMIT{the former is experimented on a simulated dataset. We finally focus on transductive PU learning problems, which are prevalent in real situations and present experimental results on two real datasets.
}

\section{Related work}\label{sec:related}

A growing body of work has focused on PU learning recently. The fact that only positive and unlabeled examples are available prevents a priori the use of supervised classification methods, which require negative examples in the training set. A first approach to overcome the lack of negative examples is to disregard unlabeled examples during training and simply learn from the positive examples, e.g., by ranking the unlabeled examples by decreasing similarity to the mean positive example \citep{Joachims97aprobabilistic} or using more advanced learning methods such as 1-class SVM \citep{Scholkopf2001Estimating,Manevitz2001One-Class,Vert2006Consistency,Bie2007Kernel-based} 

Alternatively, the problem of inductive PU learning has been studied on its own from a theoretical viewpoint \citep{Denis2005Learning,Scott2009Novelty}, and has given rise to a number of specific algorithms. Several authors have proposed two-step algorithms, heuristic in nature, which first attempt to identify negative examples in the unlabeled set, and then estimate a classifier from the positive, unlabeled and likely negative examples \citep{Manevitz2001One-Class,Liu2002Partially,Li2003Learning, Liu2003Building,Yu2004PEBL}.\OMIT{Such methods include S-EM \citep{Liu2002Partially}, Roc-SVM \citep{Li2003Learning}, PEBL \citep{Yu2004PEBL}, and several variants investigated by \citet{Liu2003Building}.} Alternatively, it was observed that directly learning to discriminate $\Pcal$ from $\Ucal$, possibly after rebalancing the misclassification costs of the two classes to account for the asymetry of the problem, leads to state-of-the-art results for inductive PU learning. This approach has been studied, with different weighting schemes, using a logistic regression or a SVM as binary classifier \citep{Liu2003Building, Lee2003Learning,Elkan2008Learning}. Inductive PU learning is also related to and has been used for novelty detection, when $\Pcal$ is interpreted as ``normal'' data and $\Ucal$ contains mostly positive examples \citep{Scott2009Novelty}, or to data retrieval from a single query, when $\Pcal$ is reduced to a singleton \citep{Shah2008SVM-HUSTLE}.

Transductive PU learning is arguably easier than inductive PU learning, since we know in advance the data to be screened for positive labels. Many semi-supervised methods have been proposed to tackle transductive learning when both positive and negative examples are known during training, including transductive SVM \citep{Joachims1999Transductive}, or many graph-based methods, reviewed by \citet{Chapelle2006Semi-Supervised}. Comparatively little effort has been devoted to the specific transductive PU learning problem, with the notable exception of \citet{Liu2002Partially}, who call the problem \emph{partially supervised classification} and proposes an iterative method to solve it, and \citet{Pelckmans2009Transductively} who formulate the problem as a combinatorial optimization problem over a graph. Finally, \citet{Sriphaew2009Cool} recently proposed a bagging approach which shares similarities with ours, but is more complex and was only tested on a specific application.

\section{Bagging for inductive PU learning}\label{sec:inductive}

Our starting point to learn a classifier in the PU learning setting is the observation that learning to discriminate positive from unlabeled samples is a good proxy to our objective, which is to discriminate positive from negative samples. Even though the unlabeled set is contaminated by hidden positive examples, it is generally admitted that its distribution contains some information which should be exploited. That is for instance, the foundation of semi-supervised methods.

 Indeed, let us assume for example that positive and negative examples are randomly generated by class-conditional distributions  $\PP_+$ and $\PP_-$ with densities $h_+$ and $h_-$. If we model unlabeled examples as randomly sampled from $\PP_+$ with probability $\gamma$ and from $\PP_-$ with probability $1-\gamma$, then the distribution of unlabeled has a density 
\begin{equation}\label{eq:gamma}
h_u = \gamma h_+ + (1-\gamma) h_-\,.
\end{equation}
Now notice that
\begin{equation}\label{eq:ratio}
\frac{h_u(x)}{h_+(x)} = \gamma + (1-\gamma) \frac{h_-(x)}{h_+(x)}\,,
\end{equation}
showing that the log-ratio between the conditional distributions of positive and unlabeled examples is monotonically increasing with the log-ratio of positive and negative examples \citep{Elkan2008Learning,Scott2009Novelty}. Hence any estimator of the conditional probability of positive vs. unlabeled data should in theory also be applicable to discriminate positive from negative examples. This is the case for example of logistic regression or some forms of SVM \citep{Steinwart2003Sparseness,Bartlett207Sparseness}. In practice it seems useful to train classifiers to discriminate $\Pcal$ from $\Ucal$ by penalizing more false negative than false positive errors, in order to account for the fact that positive examples are known to be positive, while unlabeled examples are known to contain hidden positives. Using soft margin SVM while giving high weights to false negative errors and low weights to false positive errors leads to the biased SVM approach described by \citet{Liu2003Building}, while the same strategy using a logistic regression leads to the weighted logistic regression approach of \citet{Lee2003Learning}. Both methods, tested on text categorization benchmarks, were shown to be very efficient in practice, and in particular outperformed all approaches based on heuristic identifications of true negatives in $\Ucal$.

Among the many methods for supervised binary classification which could be used to discriminate $\Pcal$ from $\Ucal$, bootstrap aggregating or ``bagging'' is an interesting candidate \citep{Breiman1996Bagging}. The idea of bagging is to estimate a series of classifiers on datasets obtained by perturbing the original training set through bootstrap resampling with replacement, and to combine these classifiers by some aggregation technique. The method is conceptually simple, can be applied in many settings, and works very well in practice \citep{Breiman2001Random,Hastie2001elements}. Bagging generally improves the performance of individual classifiers when they are not too correlated to each other, which happens in particular when the classifier is highly sensitive to small perturbations of the training set. For example, \citet{Breiman2001Random} showed that the difference between the expected mean square error (MSE) of a classifier trained on a single bootstrap sample and the MSE of the aggregated predictor increases with the variance of the classifier.

We propose that, by nature, PU learning problems have a particular structure that leads to instability of classifiers, which can be advantageously exploited by a bagging-like procedure which we now describe. Intuitively, an important source of instability in PU learning situations is the empirical contamination $\hat{\gamma}$ of $\Ucal$ with positive examples, i.e., the percentage of positive examples in $\Ucal$ which on average equals $\gamma$ in (\ref{eq:gamma}). If by chance $\Ucal$ is mostly made of negative examples, i.e., has low contamination by positive examples, then we will probably estimate a better classifier than if it contains mostly positive examples, i.e., has high contamination. Moreover, we can expect the classifiers in these different scenarii to be little correlated, since intuitively they estimate different log-ratios of conditional distribution. Hence, in addition to the ``normal'' instability of a classifier trained on a finite-size sample, which is exploited by bagging in general, we can expect an increased instability in PU learning due to the sensitivity of the classifier to the empirical contamination $\hat{\gamma}$ of $\Ucal$ in positive examples. In order to exploit this sensitivity in a bagging-like procedure, we propose to randomly subsample $\Ucal$ and train classifiers to discriminate $\Pcal$ from each subsample, before aggregating the classifiers. By subsampling $\Ucal$, we hope to vary in particular the empirical contamination between samples. This will induce a variety of situations, some lucky (small contamination), some less lucky (large contamination), which eventually will induce a large variability in the classifiers that the aggregation procedure can then exploit. 

In opposition to classical bagging, the size $K$ of the samples generated from $\Ucal$ may play an important role to balance the accuracy against the stability of individual classifiers. On the one hand, larger subsamples should lead on average to better classifiers, since any classification method generally improves on average when more training points are available. On the other hand, the empirical contamination varies more for smaller subsamples. More precisely, let us denote by $\hat{\gamma}$ the true contamination rate in $\Ucal$, that is, the true proportion of positive examples hidden in $\Ucal$. Whenever a bootstrap sample $\Ucal_t$ of size $K$ is drawn from $\Ucal$, its empirical number of positive examples is a binomial random variable $\sim B(K,\hat{\gamma})$, leading to a contamination rate $\hat{\gamma}_t$ with mean and variance:
$$\mathbb{E}(\hat{\gamma}_t)=\hat{\gamma} \textrm{ and } \mathbb{V}(\hat{\gamma}_t) = \frac{1}{K}\hat{\gamma}(1-\hat{\gamma})\,.$$
Smaller values of $K$ therefore increase the proportion of ``lucky'' subsamples, and more generally the variability of classifiers, a property which is beneficial for the aggregation procedure. Finally this suggests that the size $K$ of subsample is a parameter whose effect should be studied and perhaps tuned.

In summary, the method we propose for PU learning is presented in Algorithm \ref{algoBootInduct}. We call it bagging SVM when the classifier used to discriminate $\Pcal$ from a random subsample of $\Ucal$ is a biased SVM.
It is akin to bagging to learn to discriminate $\Pcal$ from $\Ucal$, with two important specificities. First, only $\Ucal$ is subsampled. This is to account for the fact that elements in $\Pcal$ are known to be positive, and moreover that the number of positive examples is often limited.Second, the size of subsamples is a parameter $K$ whose effect needs to be studied. If an optimal value exists, then this parameter may need to be adjusted.

The number $T$ of bootstrap samples is also a user-defined parameter. Intuitively, the larger $T$ the better, although we observed empirically little improvement for $T$ larger than $100$. Finally, although we propose to aggregate the $T$ classifiers by a simple average, other aggregation rules could easily be used. On preliminary experiments on simulated and real data, we did not observed significant differences between the simple average and majority voting, another popular aggregation method.

\section{Bagging SVM for transductive PU learning}\label{sec:transductive}

We now consider the situation where the goal is only to assign a score to the elements of $\Ucal$ reflecting our confidence that these elements belong to the positive class. \citet{Liu2002Partially} have studied this same problem which they call ``partially supervised classification''. Their proposed technique combines Naive Bayes classification and the Expectation-Maximization algorithm to iteratively produce classifiers. The training scores of these classifiers are then directly used to rank $\Ucal$. Following this approach, a straightforward solution to the transductive PU learning problem is to train any classifier to discriminate between $\Pcal$ and $\Ucal$ and to use this classifier to assign a score to the unlabeled data that were used to train it. Using SVMs this amounts to using the biased SVM training scores. We will subsequently denote this approach by transductive biased SVM. 

However, one may argue that assigning a score to an unlabeled example that has been used as negative training example is problematic. In particular, if the classifier fits too tightly to the training data, a false negative $x_i$ will hardly be given a high training score when used as a negative. In a related situation in the context of semi-supervised learning, \citet{Zhang2009Maximum} showed for example that unlabeled examples used as negative training examples tend to have underestimated scores when a SVM is trained with the classical hinge loss. More generally, most theoretical consistency properties of machine learning algorithms justify predictions on samples outside of the training set, raising questions on the use of all unlabeled samples as negative training samples at the same time.

Alternatively, the inductive bagging PU learning lends itself particularly well to the transductive setting, through the procedure described in Algorithm \ref{algoBootTransduc}. Each time a random subsample $\Ucal_t$ of $\Ucal$ is generated, a classifier is trained to discriminate $\Pcal$ from $\Ucal_t$, and used to assign a predictive score to any element of $\Ucal\setminus\Ucal_t$. At the end the score of any element $x\in\Ucal$ is obtained by aggregating the predictions of the classifiers trained on subsamples that did not contain $x$ (the counter $n(x)$ simply counts the number of such classifiers). As such, no point of $\Ucal$ is used simultaneously to train a classifier and to test it. In practice, it is useful to ensure that all elements of $\Ucal$ are not too often in $\Ucal_t$, in order to average the predictions over a sufficient number of classifiers. 
\begin{algorithm}
\caption{Inductive bagging PU learning}
\label{algoBootInduct}
\begin{algorithmic}
\STATE \small{INPUT} : $\mathcal{P}$, $\mathcal{U}$, $K =$ size of bootstrap samples, $T =$ number of bootstraps\\
\small{OUTPUT} : a function $f:\Xcal\rightarrow\mathbb{R}$
\FOR{$t=1$ to $T$}
\STATE Draw a subsample $\mathcal{U}_t$ of size $K$ from $\mathcal{U}$.\\
Train a classifier $f_t$ to discriminate $\mathcal{P}$ against $\mathcal{U}_t$.\\
\ENDFOR
\STATE Return $$f = \frac{1}{T}\sum_{t=1}^{T}f_t$$
\end{algorithmic}
\end{algorithm}

\begin{algorithm}
\caption{Transductive bagging PU learning}
\label{algoBootTransduc}
\begin{algorithmic}
\STATE \small{INPUT} : $\mathcal{P}$, $\mathcal{U}$, $K =$ size of bootstrap samples, $T =$ number of bootstraps\\
\small{OUTPUT} : a score $s:\Ucal\rightarrow\mathbb{R}$
\STATE Initialize $\forall x \in \mathcal{U}, n(x)\gets 0, f(x)\gets 0$
\FOR{$t=1$ to $T$}
\STATE Draw a bootstrap sample $\mathcal{U}_t$ of size $K$ in $\mathcal{U}$.\\
Train a classifier $f_t$ to discriminate $\mathcal{P}$ against $\mathcal{U}_t$.\\
For any $x\in {\mathcal{U}\setminus \mathcal{U}_t}$, update:
\begin{equation*}
\begin{split}
f(x) &\gets f(x) + f_t(x)\,,\\
n(x) &\gets n(x) + 1\,.
\end{split}
\end{equation*}
\ENDFOR
\STATE Return $s(x) = f(x)/n(x)$ for $x \in \mathcal{U}$
\OMIT{$$f(x) = \frac{1}{n_t(x)}\sum_{t=1}^{T} n_t(x)*f_t(x) $$}
\end{algorithmic}
\end{algorithm}

\section{Experiments}
\label{sec:res}

In this section we investigate the empirical behavior of our bagging algorithm on one simulated dataset (Section \ref{sec:toydataset}) and two real applications: text retrieval with the 20 newsgroup benchmark (Section 5.2), and reconstruction of gene regulatory networks (Section 5.3). We compare the new bagging SVM to the state-of-the-art biased SVM, and also add in the comparison for real data two one-class approaches, namely, ranking unlabeled examples by decreasing mean similarity to the positive examples (called \emph{Baseline} below), and the one-class SVM \citep{Scholkopf2001Estimating}. Both bagging and biased methods involve an SVM with asymetric penalties $C_+$ and $C_-$ for the positive and negative class, respectively. By default we always set them to ensure that the total penalty is equal for the two classes, i.e., $C_+ n_+ = C_- n_-$, where $n_+$ and $n_-$ are the number of positive and negative examples fed to the SVM, and optimized the single parameter $C=C_+ + C_-$ over a grid. We checked on all experiments that this choice was never significantly outperformed by other penalty ratio $C_+/C_-$.

\subsection{Simulated data}
\label{sec:toydataset}

A first series of experiments were conducted on simulated data to compare our bagging procedure to the biased approach in an inductive setting. We consider the simple situation where the positive examples are generated from an isotropic Gaussian distribution in $\mathbb{R}^p$ : $\Pcal \sim \mathbb{P}_{+} = \mathcal{N}(0_p, \sigma*I_p)$, with $p=50$ and $\sigma=0.6$, while the negative examples are generated from another Gaussian distribution with same isotropic covariance and a different mean, of norm $1$. We replicate the following iteration $50$ times for different values of $\gamma$~: 
\begin{itemize}
 \item Draw a sample $\Pcal$ of 5 positives examples, and a sample $\Ucal$ of 50 unlabeled examples from $\gamma*\mathbb{P}_{+} + (1-\gamma)*\mathbb{P}_{-}$.
 \item Train respectively the biased and  bagging logit (with 200 bootstraps)\footnote{The bagging logit corresponds to the procedure described above, when the classifier is a logistic regression. This is the same for the biased logit, see also\citep{Lee2003Learning}}.
\item Compare their performance on a test set of 1000 examples containing $50\%$ positives.
\end{itemize}

\begin{figure}[ht]
\centering
\includegraphics[width=7cm]{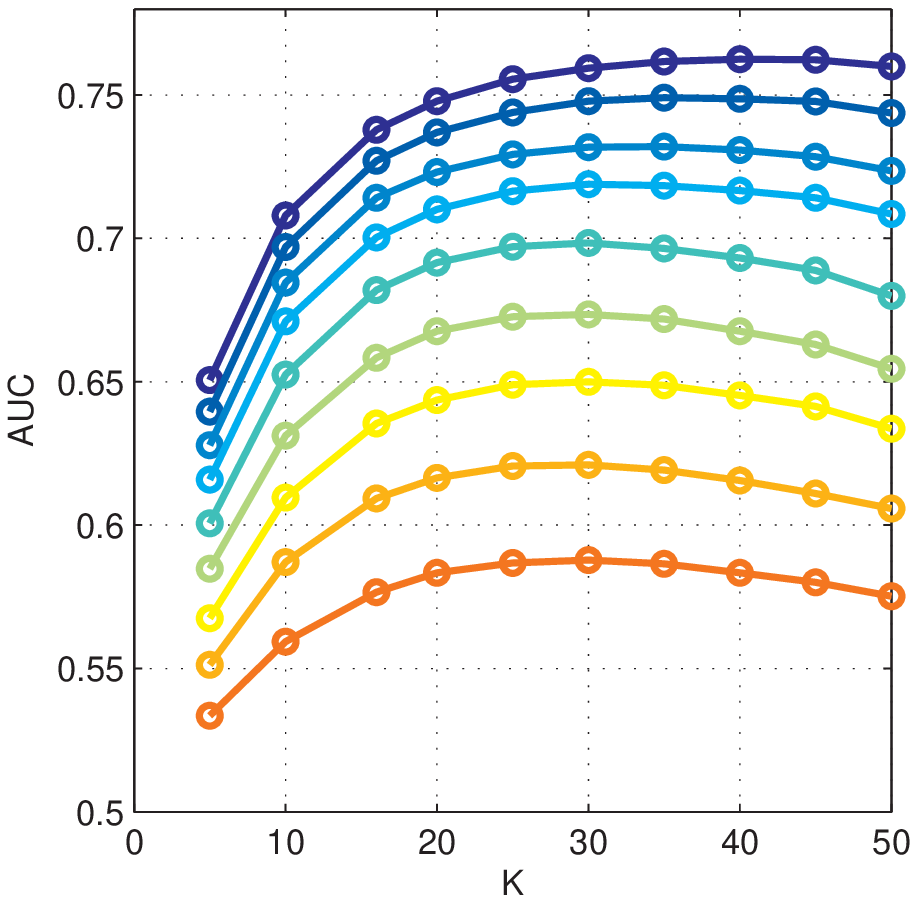}
\includegraphics[width=7cm]{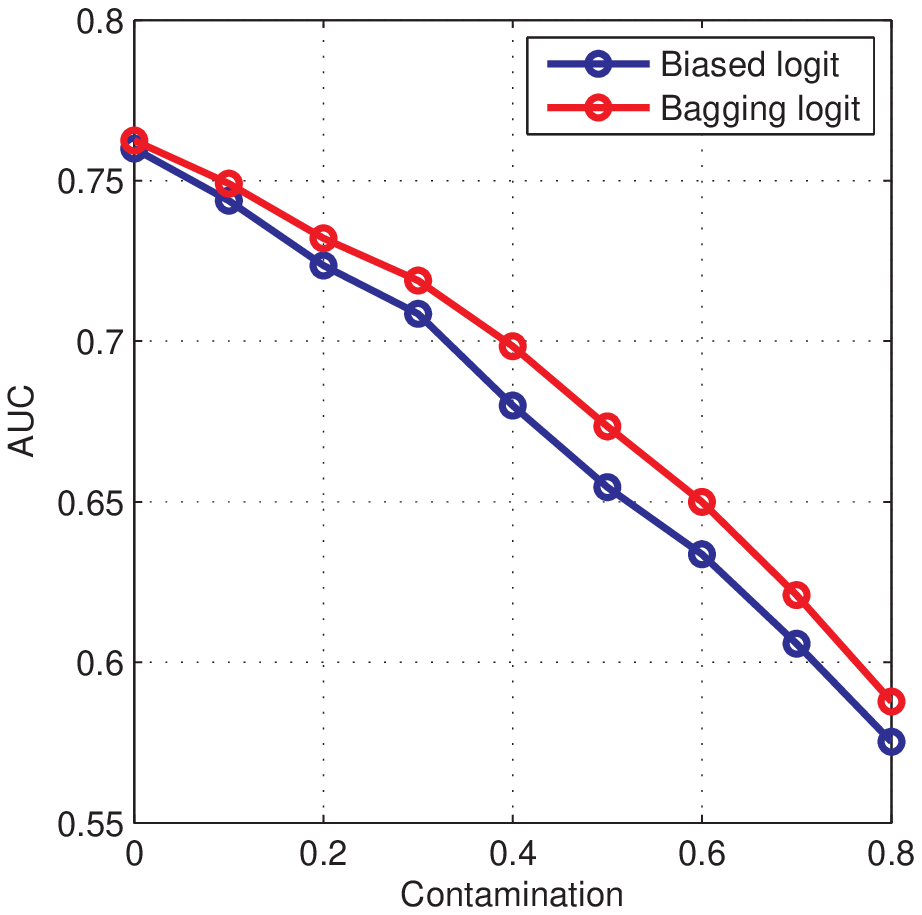}
\caption{Results on simulated data. \emph{Left:} AUC of the bagging logit as a function of $K$, the size of the bootstrap samples, on simulated data. Each curve, from top to bottom, corresponds to a contamination level $\gamma \in \{0 ; 0.1 ; 0.2 ; \ldots ; 0.8\}$. \emph{Right} Performance of two methods as a function of $\gamma$, the contamination level, on simulated data. The performance of bagging logit was taken at the optimal $K$ value.}\label{fig:toyresults}
\end{figure}

For $K$, we tested equally spaced values between 1 and 50, and we varied $\gamma$ on the interval $[0;0.9]$. The performance is measured by computing the area under the Receiving Operator Characteristic curve (AUC) on the independent test set. Figure \ref{fig:toyresults} (left) shows the performance of bagging logit for different levels of contamination of $\Ucal$, as a function of $K$, the size of the random samples. The uppermost curve thus corresponds to $\gamma=0$, i.e., the case where all unlabeled data are negative, while the bottom curve corresponds to $\gamma=0.8$, i.e., the case where $80\%$ of unlabeled data are positive. Note that $K=50$ corresponds to classical bagging on the biased logit classifier, i.e., to the case where all unlabeled examples are used to train the classifier.

We observe that in the classical setting of supervised binary classification where $\Ucal$ is not contaminated by positive samples ($\gamma=0$), the bagging procedure does not improve performance, whatever the size of the bootstrap samples. On the other hand, as contamination increases, we observe an overall decrease of the performance, confirming that the classification problem becomes more difficult when contamination increases. In addition, the bagging logit always succeeds in reaching at least the same performance for some value of $K$ below $50$, even for high rates of contamination. Figure \ref{fig:toyresults} (right) shows the evolution of AUC as $\gamma$ increases, for both methods. For the bagging logit we report the AUC reached for the best $K$ value. We see that bagging logit slightly outperforms biased logit method.

\begin{figure}[h!]
\begin{center}
\includegraphics[width=\textwidth]{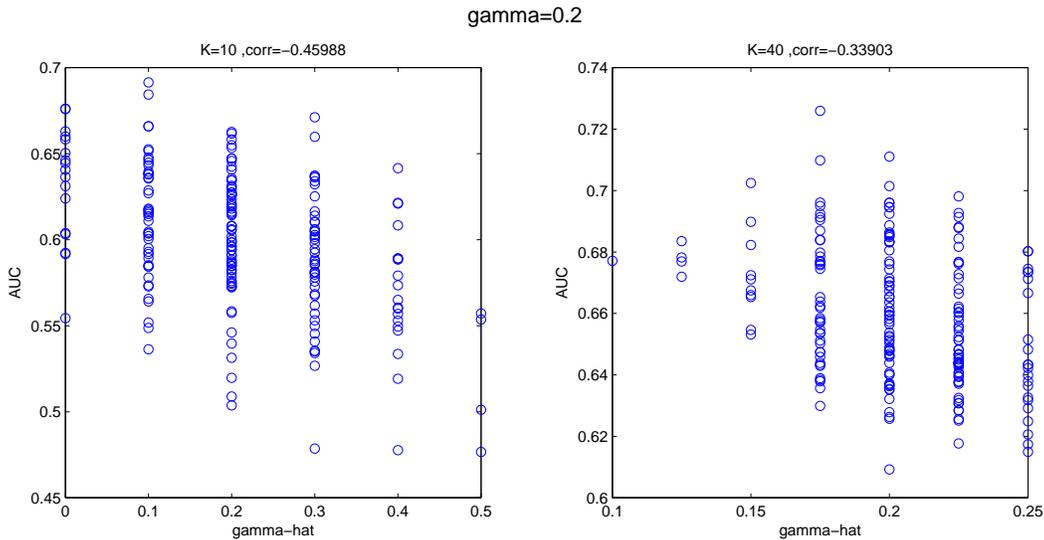} 
\end{center}
\caption{Distribution of AUC and $\hat{\gamma}$ over the 500 iterations of one bootstrap loop on the simulated dataset, $\gamma=0.2$.}
\label{fig:aucGammahat}
\end{figure}

To further illustrate the assumption that motivated bagging SVM, namely that decreasing $K$ would decrease the average performance of single classifiers but would increase their variance due to the variations in contamination, we show in Figure \ref{fig:aucGammahat} a scatter plot of the AUC of individual classifiers as a function of the empirical contamination of the bootstrap sample $\hat{\gamma}$, for two values of $K$ ($10$ and $40$). Here the mean contamination was set to $\gamma=0.2$. Obviously, the variations of $\hat{\gamma}$ are much larger for $K=10$ (between $0$ and $0.5$) than for $K=40$ (between $0.1$ and $0.25$). The correlation coefficient between $\hat{\gamma}$ and the performance (reported above each plot) is strongly negative, in particular for smaller $K$. It is quite clear that less contaminated subsamples tend to yield better classifiers, and that the variation in the contamination is an important factor to increase the variance between individual predictors, which aggregation can benefit from.

\subsection{Newsgroup dataset}
The 20 Newsgroup benchmark is widely used to test PU learning methods. The version we used is a collection of 11293 articles partitioned into 20 subsets of roughly the same size (around 500)\footnote{We used the Matlab pre-processed version available at \url{http://renatocorrea.googlepages.com/ng2011293x8165itrn.mat}}, corresponding to post articles of related interest. For each newsgroup, the positive class consists of those $\sim$500 articles known to be relevant, while the negative class is made of the remainder. After pre-processing, each article is represented by a $8165$-dimensional vector, using the TFIDF representation over a dictionnary of $8165$ words \citep{Joachims97aprobabilistic}.

To simulate a PU learning problem, we applied the following strategy. For a given newsgroup, we created a set $\Pcal$ of known positive examples by randomly selecting a given number of positive examples, while $\Ucal$ contains the non-selected positive examples and all negative examples. We varied the size $NP$ of $\Pcal$ in $\{5, 10, 20, 50, 100, 200, 300\}$ to investigate the influence of the number of known positive examples. For each newsgroup and each value of $NP$, we train all 4 methods described above (bagging SVM, biased SVM, baseline, one-class SVM) and rank the samples in $\Ucal$ by decreasing score (transductive setting). We then compute the area under the ROC curve (AUC), and average this measure over 10 replicates of each newsgroup and each value of $NP$. For bagging and biased SVM, we varied the $C$ parameter over the grid $\left[ \exp(-12:2:2)\right]$, while we vary parameter $\nu$ in $\left[ 0.1:0.1:0.9\right] $ for $1$-class SVM. We only used the linear kernel.

\begin{figure}[ht]
\centering
\includegraphics[width=16cm,height=9.5cm]{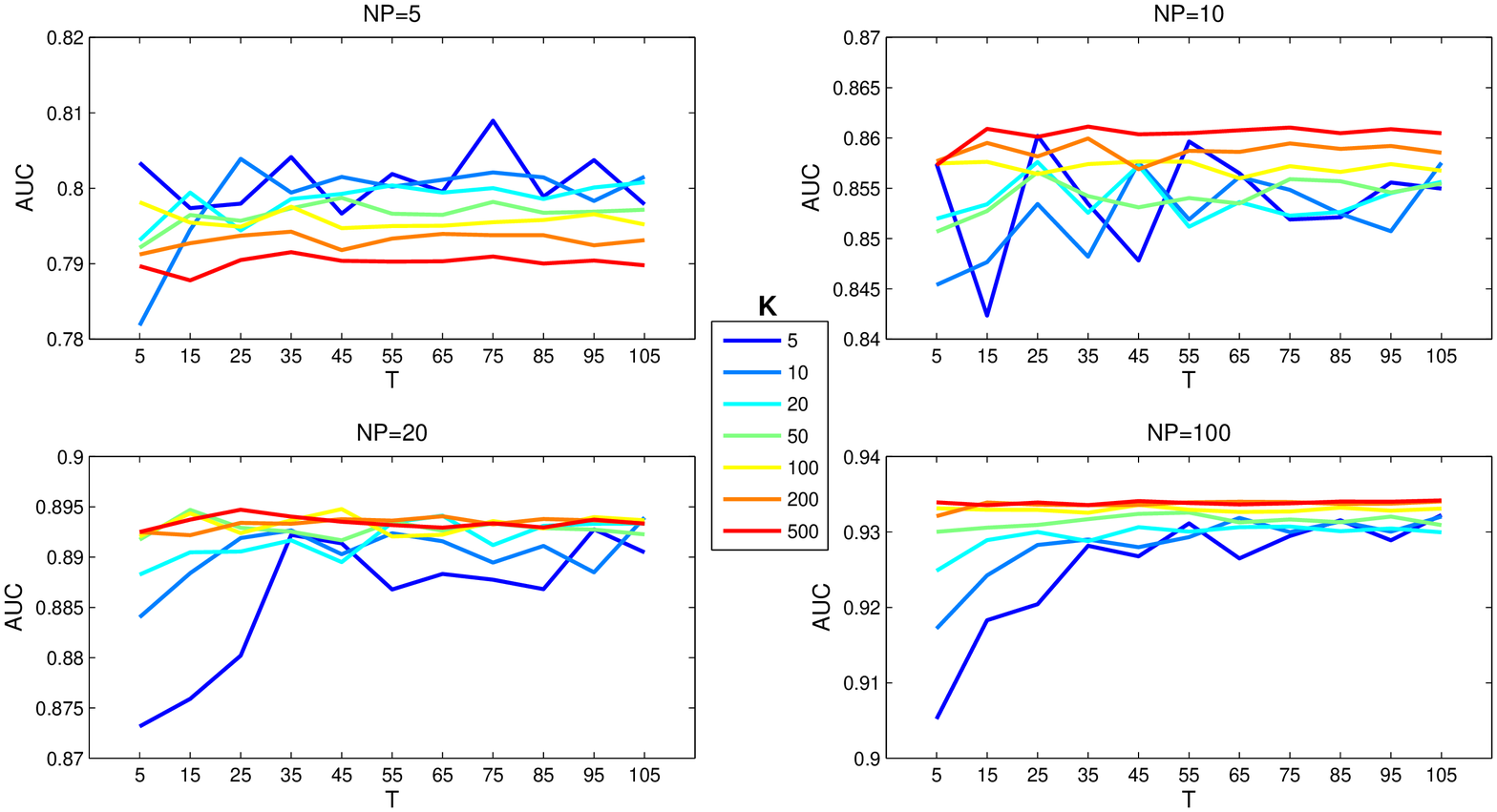}
\caption{Performance on one newsgroup as a function of the number of boostraps $T$, for different values of $NP$ and $K$.}\label{fig:bootstrapSize}
\end{figure}

\begin{figure}[ht]
\centering
\includegraphics[width=9cm,height=5.5cm]{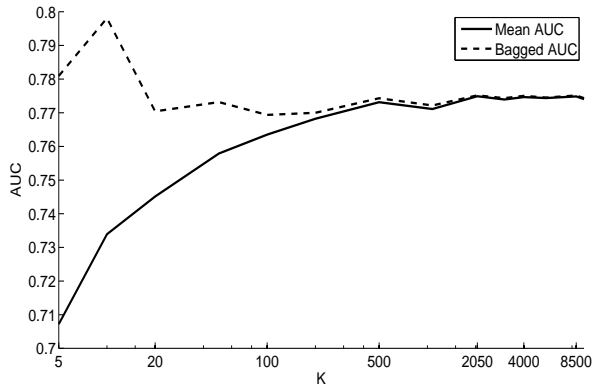}
\label{fig:baggingImprove}
\caption{Performance on one newgroup of bagging SVM (\emph{bagging AUC}) vs a SVM trained on a single bootstrap sample (\emph{mean AUC}), for different values of $K$.}
\end{figure}

We first investigated the influence of $T$. Figure \ref{fig:bootstrapSize} shows, for the first newsgroup, the performance reached as a function of $T$, for different settings in $NP$ and $K$. As expected we observe that in general the performance increases with $T$, but quickly reaches a plateau beyond which additional bootstraps do not improve performance. Overall the smaller $K$, the larger $T$ must be to reach the plateau. From these preliminary results we set $T=35$ for $K\leq 20$, and $T=10$ for $K>30$, and kept it fix for the rest of the experiments. To further clarify the benefits of bagging, we show in Figure \ref{fig:baggingImprove} the performance of the bagging SVM versus the performance of a SVM trained on a single bootstrap sample ($T=1$), for different values of $K$ and a fixed number of positives $NP=10$. We observe that, for $K$ below $200$, aggregating classifiers over several bootstrap subsamples is clearly beneficial, while for larger values of $K$ it does not really help. This is coherent with the observation that SVM usually rarely benefit from bagging: here the benefits come from our particular bagging scheme. Interestingly, we see that very good performance is reached even for small values of $K$ with the bagging.

\begin{figure}[ht]
\centering
\includegraphics[width=7.5cm, height=6cm]{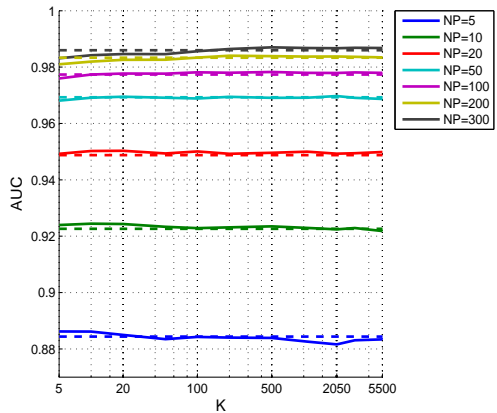}
\caption{Macro averaged performance of the bagging SVM as a function of $K$. The dashed horizontal lines show the AUC level of the biased SVM. The curves are plotted for different values of $NP$, the size of the positive set.}
\label{fig:aucBaggingK_vs_Biased}
\end{figure}

Figure \ref{fig:aucBaggingK_vs_Biased} shows the mean AUC averaged over the 10 folds and the 20 newsgroups for bagging SVM as a function of $K$, and compares it to that of the biased SVM. More precisely, each point on the curve corresponds to the performance averaged over the 20 Newsgroups after choosing a posteriori the best $C$ parameter for each newsgroup. This is equivalent to comparing optimal cases for both methods. Contrary to what we observed on simulated date, we observe that $K$ has in general very little influence on the performance. The AUC of the bagging SVM is similar to that of the biased SVM for most values of $K$, although for $NP$ larger than $50$, a slight advantage can be observed for the biased SVM over bagging SVM when $K$ is too small. We conclude that in practice, parameter $K$ may not need to be finely tuned and we advocate to keep it moderate. In all cases, $K=NP$ seems to be a safe choice for the bagging SVM.

\begin{figure}[ht]
\centering
\includegraphics[width=7.5cm, height=6cm]{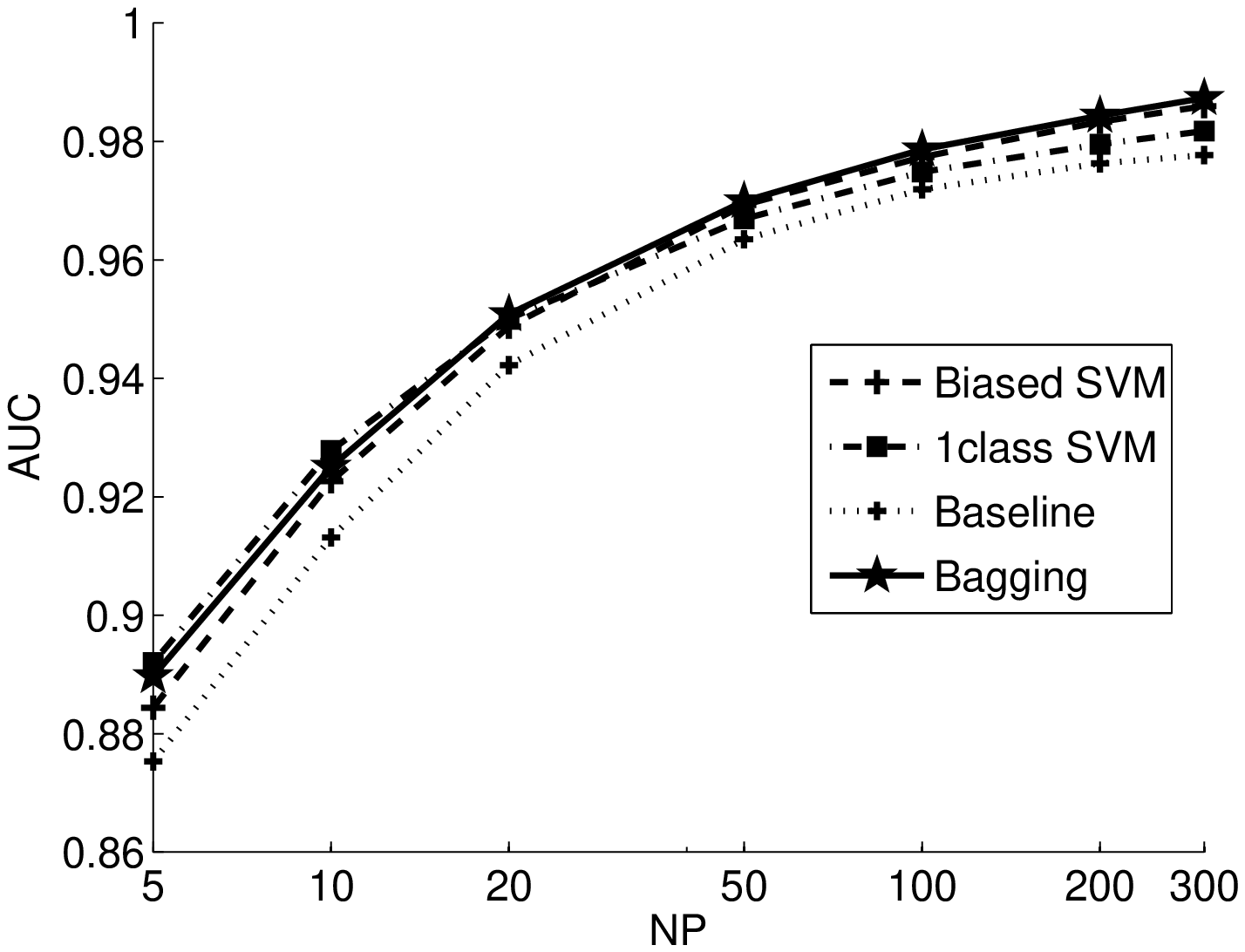}
\caption{Performance of the baseline method, the $1$-class SVM, the biased SVM and the newly proposed bagging SVM methods on the 20 Newsgroups dataset. Each curve shows how the mean AUC varies with the number of positive training examples $NP$. For each value of $NP$, the performance of bagging SVM is computed at the optimal value for $K$, as shown in Figure \ref{fig:aucBaggingK_vs_Biased}.}
\label{fig:Comp4methods}
\end{figure}

Finally, Figure \ref{fig:Comp4methods} shows the average AUC over the 20 newsgroups for all four methods, as a function of $NP$. Overall all methods are very similar, with the Baseline slightly below the others. In details, the bagging SVM curve dominates all other methods for $NP \geq 20$, while the $1$-class SVM is the one which dominates for smaller values of $NP$. Although the differences in performance are small, the bagging SVM outperforms the biased SVM significantly for $NP>20$ according to a Wilcoxon paired sample test (at $5\%$ confidence). For small values of $NP$ however, no significant difference can be proven in either way between bagging SVM and $1$-class SVM, which remains a very competitive method.

\subsection{\emph{E. coli} dataset : inference of transcriptional regulatory network}
In this section we test the different PU learning strategies on the problem of inferring the transcription regulatory network of the bacteria \emph{Escherichia coli} from gene expression data. The problem is, given a transcription factor (TF), to predict which genes it regulates. Following \citet{Mordelet2008SIRENE}, we can formulate this problem as transductive PU learning by starting from known regulated genes (considered positive examples), and looking for additional regulated genes in the bacteria's genome.

To represent the genes, we use a compendium of microarray expression profiles provided by \citet{Faith2008}, in which 4345 genes of the \emph{E. Coli} genome are represented by vectors in dimension $445$, corresponding to their expression level in 445 different experiments. We extracted the list of known regulated genes for each TF from RegulonDB \citep{Salgado2006RegulonDB}. We restrict ourselves to $31$ TFs with at least $8$ known regulated genes. 

\OMIT{For a given TF, the positive class consists in those genes which are regulated by that TF and the negative class in those which are not. The known positive set $\Pcal$ is exactly $\mathcal{C}_i$ and $\mathcal{U}$ is the remainder of the genome. The network inference problem can then be stated as follows : ``For each TF $i$ in $\mathcal{F}$, given microarray expression profiles and knowledge of $\mathcal{C}_i$, find other targets for TF $i$.'' Therefore, we see that it can be broken down into as many PU learning problems as there are transcription factors \citep{Mordelet2008SIRENE}.
}

For each TF, we ran a double 3-fold cross validation with an internal loop on each training set to select parameter $C$ of the SVM (or $\nu$ for the $1$-class SVM). Following \citet{Mordelet2008SIRENE}, we normalize the expression data to unit norm, use a Gaussian RBF kernel with $\sigma=8$, and perform a particular cross-validation scheme to ensure that operons are not split between folds. Finally, following our previous results on simulated data and the newsgroup benchmark, we test two variants of bagging SVM, setting $K$ successively to $NP$ and $5*NP$. These choices are denoted respectively by \emph{bagging1 SVM} and \emph{bagging5 SVM}.
 
\begin{figure}[ht]
\centering
\includegraphics[scale=0.5]{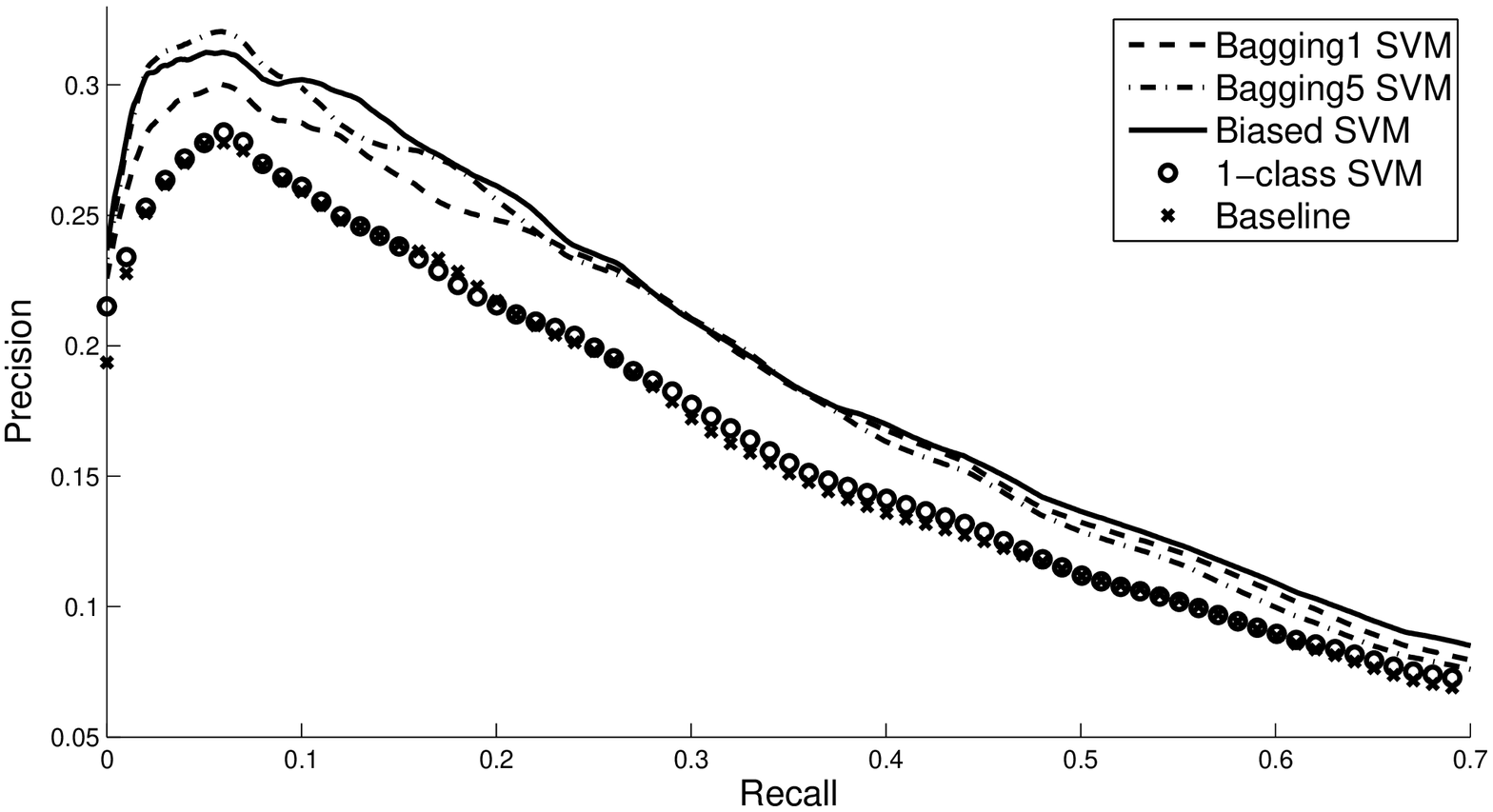}
\label{fig:PrecisionEcoli}
\caption{Precision-recall curves to compare the performance between the baggin1 SVM, the bagging5 SVM, the biased SVM, the $1$-class SVM and the baseline method.}
\end{figure}

Figure \ref{fig:PrecisionEcoli} shows the average precision/recall curves of all methods tested. Overall we observe that all three PU learning methods give significantly better results than the two methods which use only positive examples (Wilcoxon paired sample test at 5\% significance level). No significant difference was found between the three PU learning methods. This confirms again that for different values of $K$ bagging SVM matches the performance of biased SVM.
 
\section{Discussion}\label{sec:discussion}

The main contribution of this work is to propose a new method, bagging SVM, both for inductive and transductive PU learning, and to assess in detail its performance and the influence of various parameters on simulated and real data.

The motivation behind bagging SVM was to exploit an intrinsic feature of PU learning to benefit from classifier aggregation through a random subsample strategy. Indeed, by randomly sampling $K$ examples from the unlabeled examples, we can expect various contamination rates, which in turn can lead to very different single classifiers (good ones when there is little contamination, worse ones when contamination is high). Aggregating these classifiers can in turn benefit from the variations between them. This suggests that $K$ may play an important role in the final performance of bagging SVM, since it controls the trade-off between the mean and variance of individual classifiers. While we showed on simulated data that this is indeed the case, and that there can be some optimum $K$ to reach the best final accuracy, the two experiments on real data did not show any strong influence of $K$ and suggested that $K=NP$ may be a safe default choice. This is a good news since it does not increase the number of parameters to optimize for the bagging SVM and leads to balanced training sets that most classification algorithms can easily handle.

The comparison between different methods is mitigated. While bagging SVM outperforms biased SVM on simulated data, they are not significantly different on the two experiments with real data. Interestingly, while these PU learning methods were significantly better than two methods that learned from positive examples only on the gene regulatory network example, the $1$-class SVM behaved very well on the 20 newsgroup benchmark, even outperforming the PU learning methods when less than $10$ training examples were provided. Many previous works, including \citet{Liu2003Building} and \citet{Yu2004PEBL} discard $1$-class SVMs for showing a bad performance in terms of accuracy, while \citet{Manevitz2001One-Class} report the lack of robustness of this method arguing that it has proved very sensitive to changes of parameters. Our results suggest that there are cases where it remains very competitive, and that PU learning may not always be a better strategy than simply learning from positives.

Finally, the main advantage of bagging SVM over biased SVM is the computation burden, in particular when there are far more unlabeled than positive examples. Indeed, a typical algorithm, such as an SVM, trained on $N$ samples, has time complexity proportional to $N^\alpha$, with $\alpha$ between $2$ and $3$. Therefore, biased SVM has complexity proportional to $(P+U)^\alpha$ while bagging SVM's complexity is proportional $T*(P+K)^\alpha$. With the default choice $K=P$ ratio of CPU time to train the biased SVM vs the bagging SVM can therefore be expected to be $\left((P+U)/(2P)\right)^\alpha/T$. Then we conclude that bagging SVM should be faster than biased SVM as soon as $U/P > 2T^{1/\alpha}-1$. For example, taking $T=35$ and $\alpha=3$, bagging SVM should be faster than biased SVM as soon as $U/P> 6$, a situation very often encountered in practice where the ratio $U/P$ is more likely to be several orders of magnitude larger. In the two real datasets, this was always the case. Table \ref{tab:CPUtimes} reports CPU time and performance measure for training bagging SVM on the first fold of newsgroup 1 with $C$ fixed at its best value a posteriori and $NP=10$.

\begin{table}[h!]
\begin{center}
\scalebox{0.8}{
 \begin{tabular}{|c|c|c|c|c|c|c|c|}
 \cline{3-8}
 \multicolumn{2}{c|}{} & \multicolumn{3}{c|}{CPU} & \multicolumn{3}{c|}{AUC-AUP} \\
 \cline{3-8}
\multicolumn{2}{c|}{Bagging} & K=10 & K=50 & K=200 & K=10 & K=50 & \multicolumn{1}{c|}{K=200} \\
\hline
 \multirow{3}{*}{T} & 35 & 13 & 39 & 91 & 0.921-0.531 &  0.917-0.524 & 0.902-0.518 \\
\cline{2-8}
   & 50 & 18 & 54 & 127 & 0.920-0.539 & 0.914-0.522 & 0.904-0.522  \\
\cline{2-8}
   & 200 & 72 & 170 & 473 & 0.918-0.539 & 0.910-0.528 & 0.904-0.511   \\
  \hline
 \end{tabular}
 }
\end{center}
\label{tab:CPUtimes}
\caption{CPU time and performance measures for different settings of $T$ and $K$ for bagging SVM.}
\end{table}

In comparison, the biased SVM's CPU time is 227s for $\textrm{AUC}=0.932$ and $\textrm{AUP}=0.491$. This confirms that for reasonable values of $T$ and $K$, the bagging SVM is much faster than the biased SVM for a comparable performance.


\end{document}